\documentclass[10pt,conference]{IEEEtran}

\usepackage{cite}
\usepackage[pdftex]{graphicx}
\usepackage[cmex10]{amsmath}
\usepackage{amssymb}
\usepackage{amsthm}
\usepackage{array}
\usepackage{booktabs}
\usepackage{url}
\usepackage{xcolor}
\usepackage{comment}
\usepackage{tabularx}
\usepackage{makecell}

\usepackage{algorithm}
\usepackage{algpseudocode}

\graphicspath{{figs/}}
\DeclareGraphicsExtensions{.pdf,.jpeg,.png}

\newif\ifieee
 \ieeefalse

\newif\ifarxiv
\arxivtrue

\ifieee
    \ifarxiv
        \PackageError{Configuration}
        {IEEE and arXiv cannot both be enabled}
        {Set either \string\ieeefalse\space or \string\arxivfalse.}
    \fi
\fi

\ifieee
\IEEEoverridecommandlockouts
\IEEEpubid{\makebox[\columnwidth]{979-8-3195-0255-1/26/\$31.00~\copyright2026 IEEE \hfill} \hspace{\columnsep}\makebox[\columnwidth]{ }}
\fi

\ifieee
\else
    \usepackage[colorlinks=true,allcolors=black]{hyperref} %
\fi

\newif\iffinal
\finaltrue
\newcommand{\cmtid}{136}

\iffinal
\else
\usepackage[switch]{lineno}
\fi

\usepackage{transparent}
\usepackage{tikz}
\ifarxiv
    \newcommand\copyrighttext{%
      \scriptsize Accepted for presentation at SIBGRAPI 2026. The final published version will be available on IEEE~Xplore.}
    \newcommand\copyrightnotice{%
    \begin{tikzpicture}[remember picture,overlay]
    \node[anchor=south,yshift=30pt,xshift=0pt] at (current page.south) {\fbox{\transparent{0.85}\parbox{\dimexpr0.6\textwidth-\fboxsep-\fboxrule\relax}{\copyrighttext}}};
    \end{tikzpicture}%
    }
\else
\fi

\begin{document}

\title{Identity-Conditioned Latent Consistency Distillation for Face Synthesis}

\iffinal
\author{
\IEEEauthorblockN{Tiago Kienen Chaves\IEEEauthorrefmark{1}, Bernardo Biesseck\IEEEauthorrefmark{1}\IEEEauthorrefmark{2}, David Menotti\IEEEauthorrefmark{1}}
\IEEEauthorblockA{
  \IEEEauthorrefmark{1}Department of Informatics, Federal University of Paran\'{a}, Curitiba, Brazil \\
  \IEEEauthorrefmark{2}Federal Institute of Mato Grosso (IFMT), Pontes e Lacerda, MT, Brazil\\
\texttt{\{tiagokienen,menotti\}@ufpr.br bernardo.biesseck@ifmt.edu.br}}
}
\else
  \author{SIBGRAPI Paper ID: \cmtid \\[8ex]}
  \linenumbers
\fi

\maketitle

\ifarxiv
    \copyrightnotice
\else
\fi

\begin{abstract}
Diffusion models have achieved strong results in high-fidelity image synthesis, but their iterative sampling process makes large-scale generation computationally expensive.
This limitation is especially relevant when generating synthetic face datasets for face recognition, where a large number of subjects with many samples in different poses, expressions, ages, etc., are required.
In this work, we show that identity-conditioned face synthesis can be performed at a substantially lower computational cost by a latent Consistency Model with few iterations, without compromising image quality. 
For training, we distill knowledge from the foundation Diffusion Model Arc2Face (teacher) by adapting its original text-to-image pipeline to an embedding-to-face setting, replacing textual prompts with ArcFace identity embeddings. 
Our distilled model (student) generates identity-conditioned face images with an average inference time of 0.4819 seconds per image, compared with 2.102 seconds for Arc2Face, resulting in a 4.36$\times$ speed-up.
Quantitative results, based on FID scores, show that the distilled model remains competitive with Arc2Face across all evaluation protocols. On 100k generated images, it achieves near-parity on CelebA (13.921 vs. 12.928) and outperforms the teacher on WebFace42M (9.317 vs. 9.802). Further evaluations on Synth-500 and AgeDB show a moderate performance gap for the former but comparable results for the latter.
These results indicate that Arc2Face can be accelerated through task-specific latent consistency distillation while preserving high image quality for large-scale synthetic face generation.
Our proposal is publicly available at
\texttt{\url{https://github.com/UFPR-IPASP-PR/FaceRec-IdentityConsistency}}.
\end{abstract}

\IEEEpeerreviewmaketitle

\section{Introduction}

Diffusion Models (DMs) have become a dominant paradigm for image generation due to their ability to synthesize high-quality and diverse images~\cite{ho2020denoising,rombach2022high}. 
However, this quality is usually obtained through an iterative reverse process that requires multiple neural network evaluations. Even when sampling is performed in a compressed latent space, as in Latent Diffusion Models (LDMs)~\cite{rombach2022high}, inference remains significantly more expensive than a single forward pass. This cost becomes a practical limitation when the goal is not to generate a few examples, but to synthesize large datasets containing thousands or millions of images.

Synthetic face generation can support several practical applications: producing large-scale identity-balanced training sets, creating controlled benchmarks for failure analysis, generating multiple variations of the same identity under different visual conditions, enabling interactive tools for dataset design, among others.

\begin{figure}[!t]
    \centering
    \includegraphics[width=\columnwidth]{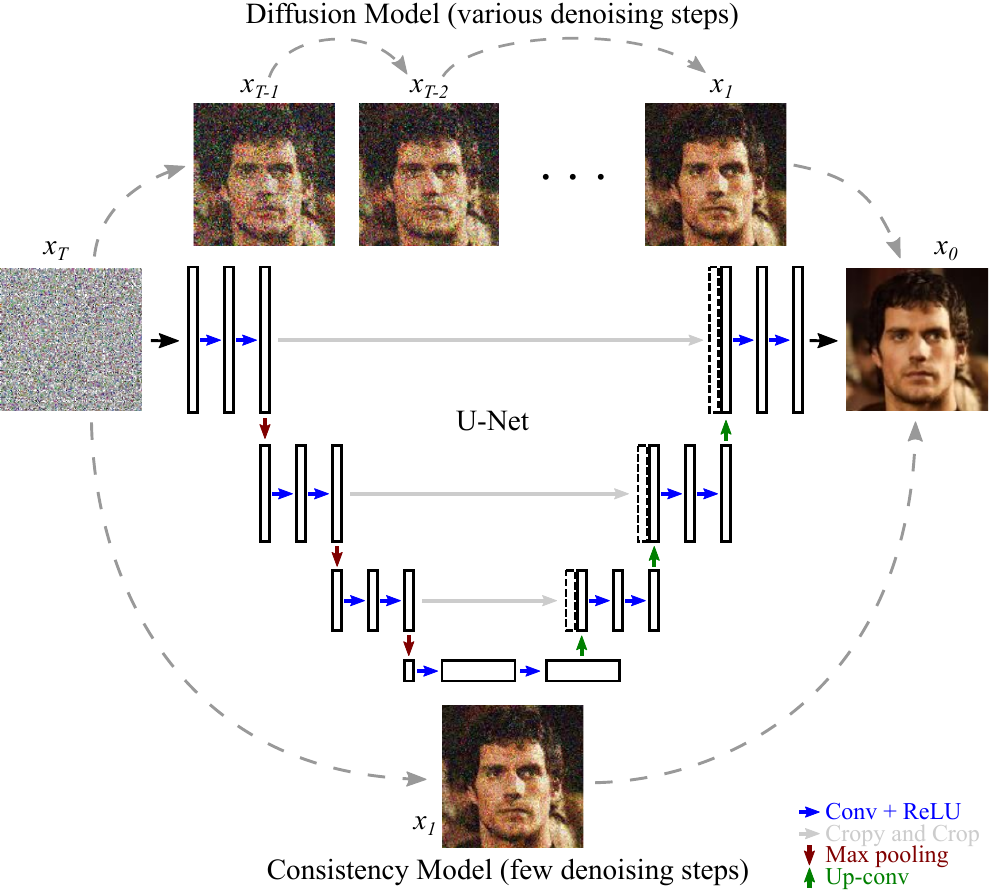}
    \vspace{-20pt}
    \caption{Illustration of the synthetic face generation processes by a Diffusion Model, such as Arc2Face (teacher), and by our distilled Consistency Model (student). While a Diffusion Model takes 25 steps to generate a high-quality image, a Consistency Model can produce equivalent results in 2 or 3 steps.}
    \label{fig_overview}
\end{figure}

In the context of face recognition, various synthetic face generation methods based on diffusion process have recently been proposed, such as Arc2Face~\cite{papantoniou2024arc2face}, VIGface~\cite{kim2025vigface}, DCFace~\cite{kim2023dcface}, Vec2Face~\cite{wu2025vec2face}, and IDiff-Face~\cite{Boutros2023IDiffFace}. These methods are capable of producing large-scale synthetic datasets but also incur high computational cost. In such a scenario, reducing generation time is not merely a convenience; it determines whether synthetic data can be produced at a scale compatible with modern face recognition experiments.

To this end, Consistency Models (CMs)~\cite{song2023consistency} were introduced as generative models capable of producing high-quality samples in one or a few steps, either by training from scratch or by knowledge distillation from pretrained diffusion models. Latent Consistency Models (LCMs)~\cite{luo2023latent} extend this idea to the latent space of pretrained Latent Diffusion Models (LDMs)~\cite{rombach2022high}, also enabling few-step high-resolution generation. LCM-LoRA further reduces the training and storage cost by learning a Low-Rank Adaptation (LoRA) module that can be plugged into diffusion models~\cite{luo2023lcmlora,hu2022lora}.

Although successfully deployed in text-to-image settings to generate stylized content (e.g., paintings, landscapes, or animated characters), these acceleration methods have only recently begun to be explored in face-related tasks, including controllable text-to-face generation and blind face restoration~\cite{shi2025expertgen,li2025interlcm}. 
However, these existing works primarily focus on image restoration or prompt matching and do not directly address the challenge of generating large-scale, identity-consistent face datasets across a variety of poses, expressions, and illumination conditions.

In this work, we bridge this gap by showing that a CM can effectively learn to generate high-quality identity-conditioned face images, comparable to those produced by DMs but in a fraction of the steps.
Our knowledge distillation experiments, using the foundational diffusion model Arc2Face~\cite{papantoniou2024arc2face} as the teacher, demonstrate that CMs can be applied to face-generation tasks that require large numbers of samples.
Our distilled model substantially reduces inference time while maintaining competitive Fréchet Inception Distance (FID scores)~\cite{heusel2017gans} relative to the diffusion teacher Arc2Face, demonstrating its potential for various applications.

\section{Related Work and Background}

In this section, we review related works and important concepts to this proposal.

\subsection{Diffusion and Latent Diffusion Models}

Denoising Diffusion Probabilistic Models (DDPMs)~\cite{ho2020denoising} generate images by learning to reverse a gradual noising process. During inference, a sample is produced through an iterative denoising trajectory, typically requiring many evaluations of a neural denoiser. Denoising Diffusion Implicit Models (DDIMs)~\cite{song2020denoising} and other numerical samplers reduce the number of steps, but generation is still fundamentally iterative. Latent Diffusion Models (LDMs)~\cite{rombach2022high} reduce the computational burden by performing denoising in the latent space of an autoencoder rather than directly in pixel space. This strategy enables high-resolution synthesis with lower memory and compute cost, and provides the foundation for many Stable Diffusion models.

\subsection{Arc2Face}

Arc2Face~\cite{papantoniou2024arc2face} is a diffusion model that produces new high-resolution face images of a subject given its face embedding. 
To ensure identity consistency, it conditions face generation on embeddings extracted by a ResNet100 (R100)~\cite{He_2016_CVPR}, pretrained on the  Webface42M~\cite{zhu2021webface260m} dataset via the ArcFace~\cite{deng2019arcface} loss function.
Specifically, the framework is built upon a latent diffusion architecture, replacing generic textual conditioning with an identity-conditioning pathway. 
While this design makes Arc2Face attractive for generating synthetic datasets for face recognition, its sampling cost scales heavily with the number of diffusion steps.

\subsection{Consistency Models and Latent Consistency Models}

Consistency models learn mappings that send noisy samples at different noise levels to a consistent point on the data manifold~\cite{song2023consistency}. They can be trained directly from data or distilled from pretrained diffusion models.
In the distillation setting, the model learns to approximate the solution of the probability flow (PF) ordinary differential equation (ODE) associated with the teacher diffusion model. 
Latent Consistency Models apply this principle in the latent space of pretrained latent diffusion models, allowing high-quality generation with very few inference steps~\cite{luo2023latent}. 
The LCM training objective combines a skipping-step strategy with guided distillation, allowing a student model to approximate the teacher's reverse process over larger jumps in the denoising trajectory.

\subsection{LCM-LoRA and Adapter-Based Acceleration}

LCM-LoRA extends latent consistency distillation by training a LoRA adapter rather than the full model~\cite{luo2023lcmlora}. LoRA reduces the number of trainable parameters by representing weight updates through low-rank matrices~\cite{hu2022lora}. This makes LCM-LoRA attractive when the goal is to accelerate generic Stable-Diffusion-based models with low training and storage overhead. However, adapter-based acceleration is not necessarily optimal for specialized settings in which the conditioning signal and evaluation criteria differ from generic text-to-image generation. In identity-conditioned face synthesis, the central requirement is not only visual realism but also preservation of the identity encoded in the input ArcFace embedding.

\subsection{Consistency-Based Face Generation and Restoration}

Although consistency-based methods have been explored in face-related contexts, most existing works do not address embedding-to-face distillation. ExpertGen~\cite{shi2025expertgen}, for example, studies controllable text-to-face generation with training-free expert guidance, while InterLCM~\cite{li2025interlcm} uses latent consistency models for blind face restoration. 
These works indicate that consistency-based models can be useful in facial domains. However, they differ from our objective: distilling an
ArcFace-embedding-conditioned face generator into a fast model for identity-conditioned synthetic data generation.

\section{Methodology}

The goal of our work is to investigate the potential of Consistency Models to generate realistic, identity-consistent face images of comparable quality to those produced by equivalent Diffusion Models, with fewer inference steps. To do this, we distill knowledge from the teacher DM Arc2Face, denoted by $\epsilon_{\psi}$, to train our student Consistency Model architecture~\cite{song2023consistency}. Our approach follows the Latent Consistency Distillation (LCD)~\cite{luo2023latent} framework while replacing text conditioning with Arc2Face identity conditioning. This section describes the identity-conditioned latent representation, the consistency mapping objective, the EMA-based distillation loss, and the complete training procedure summarized in Algorithm~\ref{alg:ic_lcd}.

Let $I$ be an input real facial image, and $e \in \mathbb{R}^{512}$ is its normalized face identity embedding extracted by an R100/ArcFace face recognition model, which is then mapped via a CLIP~\cite{pmlr_v139_radford21a} text encoder to match the dimensions expected by the cross-attention layers of the Stable Diffusion U-Net~\cite{ronneberger2015unet} architecture:
\begin{equation}
c = \text{Project}(e).
\end{equation}

The frozen Arc2Face teacher network $\epsilon_{\psi}$ operates within a latent space generated by a Variational Autoencoder (VAE) $\mathcal{E}$, with the latent image $z_0 = \mathcal{E}(I)$. The forward diffusion process adds Gaussian noise to the latent vector over a discrete timeline, yielding a sample $z_t$ at time step $t$. In each iteration, the teacher model yields a noise prediction $\epsilon_{\psi}(z_t, c, t)$.

To enforce powerful classifier-free guidance during distillation, we define the guided noise prediction sequence using the Imagen-style formulation. Given a guidance parameter $w \sim [\omega_{\text{min}}, \omega_{\text{max}}]$, the guided teacher prediction is defined linearly as:
\begin{equation}
\tilde{\epsilon}*{\psi}(z_t, c, \emptyset, w, t) = \epsilon*{\psi}(z_t, \emptyset, t) + w \cdot \left[ \epsilon_{\psi}(z_t, c, t) - \epsilon_{\psi}(z_t, \emptyset, t) \right],
\end{equation}
\noindent where $\emptyset$ represents the unconditional null token or empty text embedding sequence used to simulate unguided generation.

\subsection{Consistency Mapping and the Consistency Objective}
\label{sec:consistencymapobj}

Our student consistency network, parameterized by $\theta$, aims to model a consistent function $f_\theta: (z_t, w, c, t) \rightarrow z_0$ that maps any point along an ODE trajectory directly to its origin. Following the boundary conditions of consistency models, the model is formulated as:
\begin{equation}
f_\theta(z_t, w, c, t) = c_{\text{skip}}(t) z_t + c_{\text{out}}(t) F_\theta(z_t, w, c, t),
\end{equation}

\noindent  where $c_{\text{skip}}(t)$ and $c_{\text{out}}(t)$ are analytical boundary functions ensuring $f_\theta(z_0, w, c, 0) = z_0$, and $F_\theta$ is a trainable UNet architecture parameterized by online weights $\theta$. Unlike standard LCD, our student architecture explicitly ingests the conditional vector $w$ via a sinusoidal Fourier embedding, mapping the guidance scale into a high-dimensional vector injected directly into the network blocks as a temporal condition.

Given a sub-sampled discrete time sequence path $\tau = [t_1, t_2, \dots, t_N]$ extracted from a total of $T$ diffusion steps (e.g., $N=50$, $T=1000$), we sample adjacent indices $(n, n+k)$ to compute the trajectory matching objective, where $k=1$ in our implementation.

Although $k=1$ in the reduced sequence, each transition still corresponds to a larger jump in the original diffusion timeline. For example, with $N=50$ reduced steps extracted from $T=1000$ diffusion steps, each adjacent transition in $\tau$ skips approximately $T/N$ original timesteps. For a given latent sample $z_{t_{n+k}}$, a single-step DDIM solver estimation $\hat{z}*{t_n}^{\Psi}$ is evaluated using the teacher's CFG output:
\begin{equation}
\hat{z}*{t_n}^{\Psi} = \text{DDIM}\left(z_{t_{n+k}}, \tilde{\epsilon}*{\psi}(z*{t_{n+k}}, c, \emptyset, w, t_{n+k}), t_{n+k}, t_n\right).
\end{equation}

\subsection{Distillation Loss with Exponential Moving Average (EMA) Targets}

To enforce training stability and protect the identity-conditioned manifold from collapsing, we follow Consistency Models and Latent Consistency Distillation~\cite{song2023consistency,luo2023latent} and employ an Exponential Moving Average (EMA) target student network parameterized by $\theta^{-}$.

The online student parameters $\theta$ are updated by minimizing the distance between the online consistency mapping at the nosier step and the target consistency mapping at the resolved DDIM step.

We optimize the online parameters $\theta$ using either a Mean Squared Error ($\mathcal{L}_2$), or a Huber loss function:

\begin{equation}
\begin{aligned}
\mathcal{L}_{\text{IC-LCD}}(\theta; \theta^{-})
  &= \mathbb{E}_{z_0, c, w, n} \Bigl[
      \mathcal{D}_{\text{Huber}}\bigl(
        f_\theta(z_{t_{n+k}}, w, c, t_{n+k}),
        \\
        & \qquad\qquad\qquad\qquad
        f_{\theta^{-}}(\hat{z}_{t_n}^{\Psi}, w, c, t_n)
      \bigr)
    \Bigr].
\end{aligned}
\end{equation}

Following each optimization step, the target network parameters $\theta^{-}$ are slowly updated via a momentum parameter $\mu$, i.e., $\theta^{-} \leftarrow \mu \theta^{-} + (1 - \mu) \theta.$

\begin{algorithm}[!t]
\caption{Identity-Conditioned Latent Consistency Distillation (IC-LCD)}
\label{alg:ic_lcd}
\begin{algorithmic}[1]
\Require Pre-trained Arc2Face Teacher U-Net $\epsilon_{\psi}$
\Require Pre-trained VAE $\mathcal{E}$
\Require Pre-trained ResNet100+ArcFace
\Require Time schedule $\tau = [t_1, t_2, \dots, t_N]$
\Require Guidance boundaries $[\omega_{\text{min}}, \omega_{\text{max}}]$
\Require Huber loss function $\mathcal{D}*{\text{Huber}}$
\Require Initial Student parameters $\theta$
\Require Target Student parameters $\theta^{-} \leftarrow \theta$,
\Require Learning rate $\eta$
\Require EMA decay $\mu$
\Repeat
\State $(e, z_0) \sim \mathcal{D}$
\State $c \leftarrow \text{Project}(e)$
\State $\emptyset \leftarrow \text{Project}(\mathbf{0})$
\State $n \sim \mathcal{U}(1, N-1)$
\State $t_n \leftarrow \tau[n]$ and $t*{n+k} \leftarrow \tau[n+1]$
\State $w \sim \mathcal{U}(\omega_{\text{min}}, \omega_{\text{max}})$
\State $\epsilon \sim \mathcal{N}(\mathbf{0}, \mathbf{I})$
\State $z_{t_{n+k}} \leftarrow \alpha_{t_{n+k}} z_0 + \sigma_{t_{n+k}} \epsilon$
\State $\epsilon_{\text{uncond}} \leftarrow \epsilon_{\psi}(z_{t_{n+k}}, \emptyset, t_{n+k})$
\State $\epsilon_{\text{cond}} \leftarrow \epsilon_{\psi}(z_{t_{n+k}}, c, t_{n+k})$
\State $\tilde{\epsilon} \leftarrow \epsilon_{\text{uncond}} + w \cdot (\epsilon_{\text{cond}} - \epsilon_{\text{uncond}})$

\State $\hat{z}_{t_n}^{\Psi} \leftarrow \text{DDIMSolver}(z_{t_{n+k}}, \tilde{\epsilon}, t_{n+k}, t_n)$

\State $e_w \leftarrow \text{GuidanceEmbedding}(w)$

\State $\hat{x}_{\theta} \leftarrow f_\theta(z_{t_{n+k}}, e_w, c, t_{n+k})$
\State $\hat{x}_{\theta^{-}} \leftarrow f_{\theta^{-}}(\hat{z}_{t_n}^{\Psi}, e_w, c, t_n)$
\State $\mathcal{L} \leftarrow \mathcal{D}_{\text{Huber}}(\hat{x}_{\theta}, \, \hat{x}_{\theta^{-}})$
\State $\theta \leftarrow \theta - \eta \nabla_\theta \mathcal{L}$.

\State $\theta^{-} \leftarrow \mu \theta^{-} + (1 - \mu) \theta$
```

\Until{converged}
\end{algorithmic}
\end{algorithm}

Algorithm~\ref{alg:ic_lcd} presents all distillation steps in detail.

\section{Experiments}

This section presents our experimental setup, training and testing datasets, obtained results, and qualitative analysis.

\subsection{Training Data and Cache Precomputation}

Due to time constraints, we used 150k face samples from the dataset WebFace4M~\cite{zhu2021webface260m}, which contains approximately 4 million face images across 200k identities and is a subset of WebFace260M~\cite{zhu2021webface260m}. All samples in that dataset are originally aligned and cropped to $112 \times 112$ pixels using five landmarks: left eye, right eye, nose tip, left mouth, and right mouth. To ensure data dimensions were compatible with the teacher model Arc2Face, face images were resized to $512\times512$ pixels and their values were normalized from $[0,255]$ to $[-1,1]$.

To avoid repeatedly loading images, extracting identity embeddings, and encoding images into latents during training, we precompute and save to disk the required tensors before distillation. The resulting cache is saved in tensor shards containing:

\begin{itemize}
    \item $e$: ArcFace identity embeddings with shape $[N,512]$ computed from original $112\times112$ pixel images;
    \item $z_0$: VAE latents with shape $[N,4,64,64]$ from $512 \times 512$ pixel images;
\end{itemize}

Our model was trained for 8 epochs with batch=2 using the optimizer AdamW with learning rate of 1e-5 on one GPU NVIDIA GeForce RTX 3090, which took approximately 28 hours.

After training, we compare our distilled consistency model with the original Arc2Face teacher. Both models receive R100/ArcFace identity embeddings as conditioning and generate new synthetic face images of such a subject at $512\times512$ pixels. The main evaluation focuses on three aspects: computational cost, distributional image quality measured by Fréchet Inception Distance (FID)~\cite{heusel2017gans}, and qualitative comparison.

FID is computed between two sets of images and measures the distance between their data distributions. Let $x^r$ and $x^s$ denote the 2048-dimensional features extracted from real and synthetic image samples, respectively, with a pretrained Inception-V3~\cite{szegedy2016rethinking} network from its intermediate activations layer \textit{pool3}. Real means and covariances are computed as $\mu_r = \frac{1}{N}\sum_{i=1}^{N}{f(x_i^r)}$ and $\Sigma_r = \frac{1}{N}\sum_{i=1}^{N}{(f(x_i^r)-\mu_r)(f(x_i^r)-\mu_r)^T}$, respectively. Both $\mu_s$ and $\Sigma_s$ are computed analogously from the synthetic images. The FID is then defined as
\begin{multline}
FID(\mu_r, \Sigma_r, \mu_s, \Sigma_s) = \\
\sqrt{
\|\mu_r - \mu_s\|_2^2
+
\operatorname{Tr}
\left(
\Sigma_r + \Sigma_s - 2(\Sigma_r\Sigma_s)^{1/2}
\right)
}.
\end{multline}

A lower FID indicates that the distribution of generated images is closer to the reference distribution in the metric's feature space.

We evaluate two settings using FID. First, we generate 100k images from a random subset of identities from WebFace4M~\cite{zhu2021webface260m}, which we compare to CelebA~\cite{liu2015deep} and another disjoint subset of 100k images from WebFace42M. This setting evaluates large-scale distributional quality both against a general face dataset, CelebA, and against a reference domain closer to the training and face-recognition data, WebFace42M. Second, we repeat the Arc2Face evaluation protocol, in which five synthetic images are generated for each synthetic identity in the datasets Synth-500~\cite{papantoniou2024arc2face} and AgeDB~\cite{moschoglou2017agedb}. This second setting is included to make our results directly comparable with the published Arc2Face protocol and to evaluate the model on both synthetic identities and real face images outside the WebFace training subset. Then, FID is computed between the corresponding samples.

\subsection{Quantitative Results}

Table~\ref{tab:fid_all} presents the FID scores for all evaluated settings. Compared to CelebA~\cite{liu2015deep}, the consistency model achieves a slightly higher FID than the teacher Arc2Face, increasing from 12.928 to 13.921, which corresponds to an absolute difference of 0.993, or approximately 7.68\%. Compared to WebFace42M~\cite{zhu2021webface260m}, the consistency model achieves a lower FID than the teacher, reducing it from 9.802 to 9.317, an improvement of approximately 4.95\%. Since WebFace4M/WebFace42M are closer to the training domain, this suggests that the distilled model preserves the distributional characteristics of the teacher particularly well in the face recognition data domain.

\begin{figure}[!t]
    \centering
    \includegraphics[width=\columnwidth]{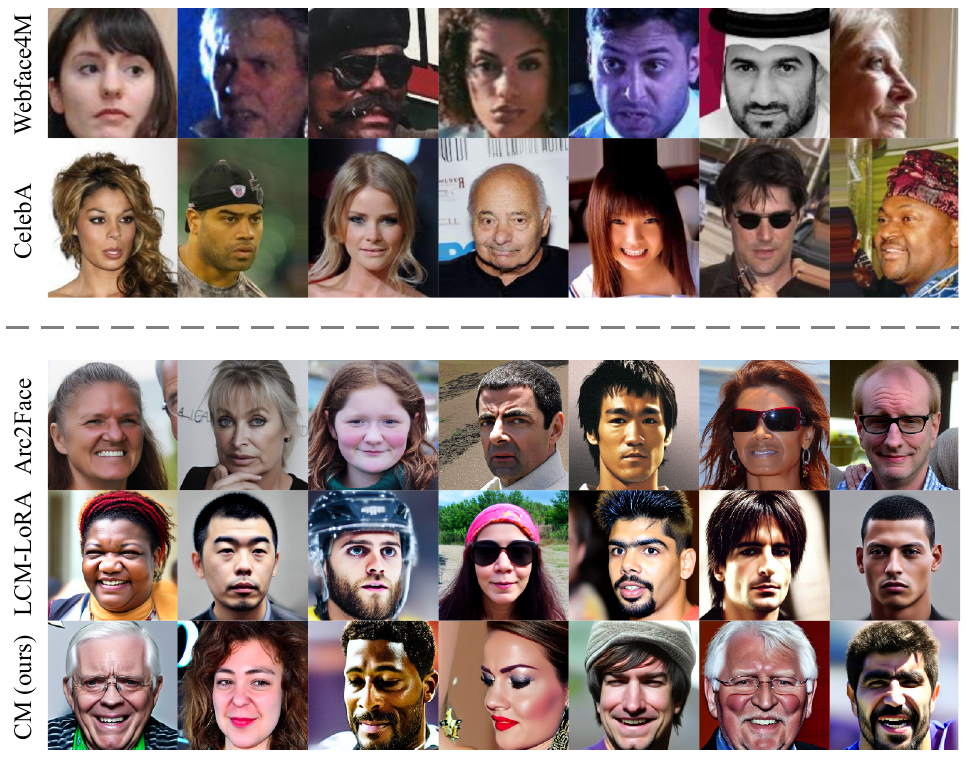}
    \vspace{-20pt}
    \caption{Real and Synthetic face images. The first two rows show samples from the real datasets Webface4M and CelebA, used in this work as quality reference. The last three rows show synthetic images generated by the teacher Diffusion Model Arc2Face, LCM-LoRA, and our distilled Consistency Model (CM) from real faces of the dataset Webface4M. Each image belongs to a distinct subject and shows the ability of all methods to generate faces of various ethnic groups.}
    \label{fig_generated_images_webface4m}
\end{figure}

Figure~\ref{fig_generated_images_webface4m} qualitatively illustrates samples from the real reference datasets and from the generated images used in the 100k-image evaluation.

\begin{table}[!t]
    \centering
    \caption{FID scores across all evaluated settings. Lower is better.}
    \vspace{-5pt}
    \label{tab:fid_all}
    \resizebox{\columnwidth}{!}{
        \begin{tabular}{lcccc}
            \toprule
            Dataset & Arc2Face & LCM-LoRA & CM (ours) & Diff. to Arc2Face \\
            \midrule
            CelebA     & 12.928 & 11.945  & 13.921  & +0.993 \\
            WebFace42M &  9.802 & 13.832  &  9.317  & $-$0.485 \\
            \midrule
            Synth-500  &  5.673 & 11.4902 &  8.7039 & +3.0309 \\
            AgeDB      &  6.628 & 13.0664 &  6.9720 & +0.3440 \\
            \bottomrule
        \end{tabular}
    }
\end{table}

These FID results indicate that the few-step distilled model does not suffer a large distributional degradation relative to the diffusion teacher. The CelebA result shows a small loss relative to Arc2Face, while the WebFace42M result shows that the consistency model can even outperform the teacher on FID in a reference domain closer to the training data.

\begin{figure}[!t]
    \centering
    \includegraphics[width=\columnwidth]{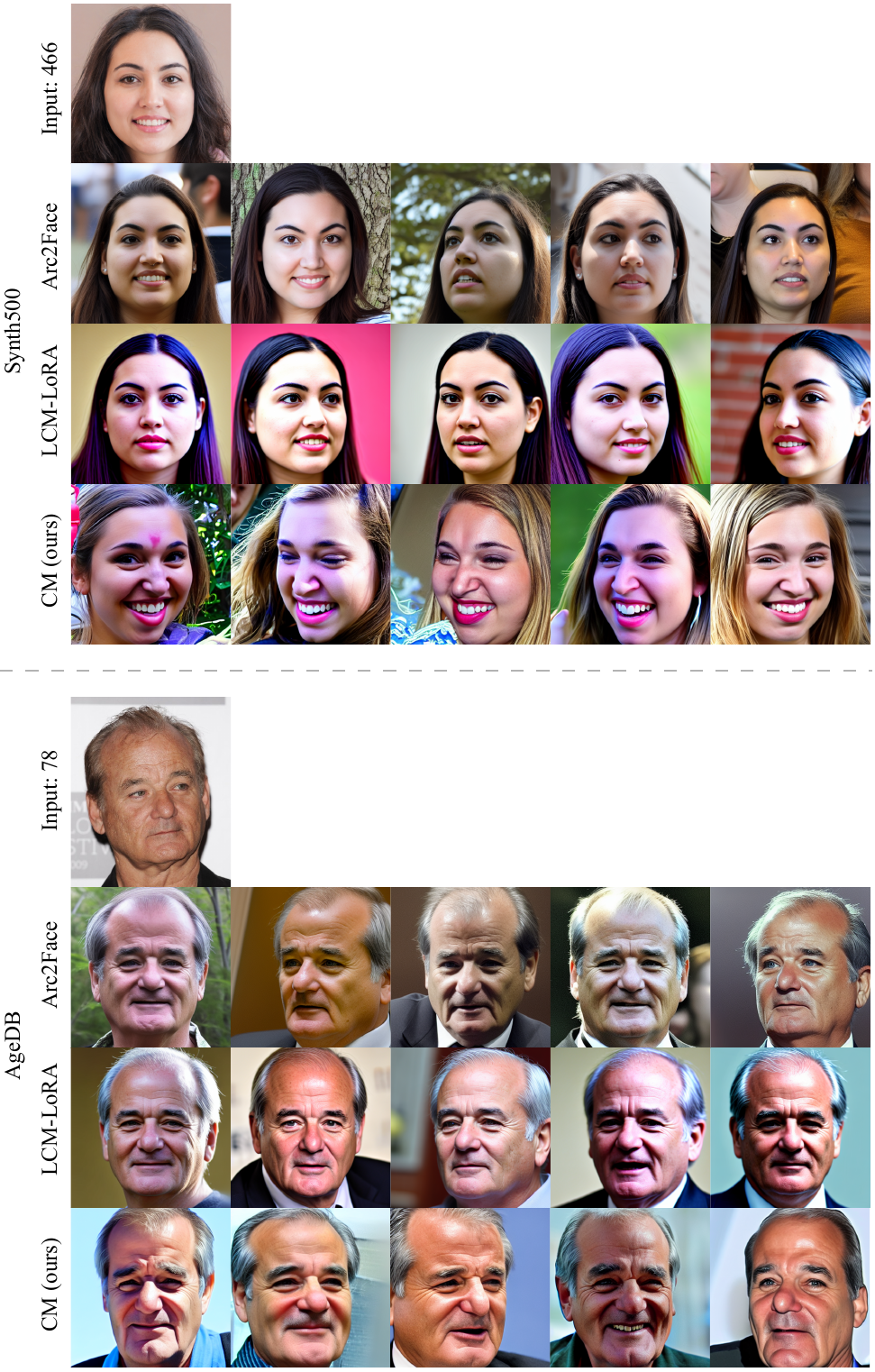}
    \vspace{-20pt}
    \caption{New generated face images from the datasets Synth-500~\cite{papantoniou2024arc2face} and AgeDB~\cite{moschoglou2017agedb}, using three different models: Diffusion Model Arc2Face (teacher), LCM-LoRA, and our Consistency Model (CM).}
    \label{fig_generated_images_synth500_agedb}
\end{figure}

For the Arc2Face protocol, the consistency model obtains a higher FID than Arc2Face on Synth-500, indicating a larger gap in this setting. On AgeDB, the difference is smaller, with the proposed model obtaining 6.9720 compared with 6.628 for Arc2Face. Figure~\ref{fig_generated_images_synth500_agedb} shows qualitative samples generated under this protocol.

The Synth-500 result shows that the student still has limitations and may not fully reproduce the teacher's distribution in all evaluation settings. One possible explanation is that Synth-500 is itself composed of synthetic identities, which may differ from the real-face distribution used during cache construction and distillation. This domain mismatch can make the distilled model less accurate in reproducing the teacher behavior under this protocol. However, the AgeDB result is close to the teacher's, suggesting that the consistency model retains much of the teacher's generation quality on real face images while substantially reducing computational cost.

\subsection{Computational Cost Comparison}

Table~\ref{tab:time} reports the average generation time per image. The distilled model reduces the average time from 2.102 seconds to 0.4819 seconds, corresponding to a 4.36$\times$ speed-up and a 77.1\% reduction in per-image generation time. Because absolute runtime depends on hardware and implementation details, the most relevant observation is the relative reduction obtained under the same evaluation setup.

\begin{table}[!b]
    \centering
    \caption{Average generation time per image.}
    \vspace{-5pt}
    \label{tab:time}
    \begin{tabular}{p{4cm}cr}
        \toprule
        Model & Time/image (s) & Relative speed \\
        \midrule
        Arc2Face diffusion teacher & 2.1020 &  1.00$\times$ \\
        LCM-LoRA                   & 0.1780 & 11.80$\times$ \\
        Proposed consistency model & 0.4819 &  4.36$\times$ \\
        \bottomrule
    \end{tabular}
\end{table}

This result is important for practical face synthesis. For example, generating 500k images would require approximately 292 GPU-hours at 2.102 seconds per image, but approximately 67 GPU-hours at 0.4819 seconds per image, assuming the same single-image throughput. Thus, the distilled model makes large-scale synthetic dataset generation substantially more feasible.

\subsection{Qualitative Results}

Qualitatively, the distilled model successfully generates recognizable face images from ArcFace embeddings. The generated samples preserve the central behavior expected from Arc2Face: the input is not a text prompt but a face recognition vector, and the output is a plausible face consistent with that vector. Visual inspection in Figures~\ref{fig_generated_images_webface4m} and~\ref{fig_generated_images_synth500_agedb} shows that the student can synthesize diverse faces and does not collapse to a single appearance pattern.

\subsection{Analysis}

The results support the hypothesis that identity-conditioned face synthesis can be accelerated through latent consistency distillation. The most direct gain is computational: the proposed model reduces generation time by more than four times. This has immediate practical consequences for synthetic data generation, where the total cost scales linearly with the number of generated images.

The FID results also suggest that the distilled model remains useful for large-scale generation. Although it is not uniformly better than Arc2Face, it remains close to the teacher in most settings and even obtains a lower FID against WebFace42M. This is relevant because WebFace-style data is directly connected to face recognition research. Therefore, for applications where a small quality trade-off is acceptable in exchange for a large reduction in cost, the proposed model is a practical alternative.

A full pretrained consistency model also has advantages over a generic LCM-LoRA adapter in this setting. LCM-LoRA is attractive because it is lightweight and plug-and-play, but it is designed primarily as a universal acceleration module for Stable-Diffusion-like models. Our setting is more specialized: the conditioning signal is an ArcFace embedding, and the key requirement is identity preservation rather than only prompt-image alignment or visual appeal. By training the student directly on Arc2Face identity-conditioned trajectories, the model can learn the task-specific relationship between identity embeddings and generated faces.

At the same time, this work does not eliminate the usefulness of adapters. LCM-LoRA remains a strong baseline for future comparison because it requires fewer trainable parameters and is easier to train. The contribution of this paper is complementary: it investigates whether a full Arc2Face-distilled consistency model can serve as a specialized generator for identity-conditioned face synthesis.

\section{Conclusion}

This paper presented a latent-consistency distillation approach to accelerate Arc2Face, an identity-conditioned face generator based on ArcFace embeddings. 
We adapted a text-to-image latent consistency distillation pipeline to the embedding-to-face setting, precomputed ArcFace identity embeddings and VAE latents for 150k WebFace4M identity samples, and trained a student model initialized from the Arc2Face teacher. The resulting model generates face images in 0.4819 seconds on average, compared with 2.1020 seconds for the diffusion teacher, achieving a 4.36$\times$ speed-up.

Quantitative results showed that the distilled model is competitive with Arc2Face. On 100k generated images, the proposed model achieves FID scores of 13.9210 on CelebA and 9.3170 on WebFace42M, compared with 12.9280 and 9.8020 for Arc2Face, respectively. In the Arc2Face protocol, the proposed model achieves 8.7039 on Synth-500 and 6.9720 on AgeDB, showing a larger gap on Synth-500 (+3.0309) but near-parity on AgeDB (+0.3440).

The main limitation of the current version is that evaluation is still centered on FID and runtime. Future work should include identity-specific metrics such as cosine similarity between conditioning embeddings and generated images, face detection rate, intra-identity diversity, inter-identity separability, and downstream face recognition performance after training with the generated data. Future ablations should evaluate the effect of the number of inference steps, guidance scale, training length, cache size, and the comparison between full-model distillation and LCM-LoRA adapters. Because we employed an embedding-to-face schema, our model is limited to ArcFace embeddings and will not work properly directly with text embeddings without retraining. Finally, because face synthesis can be misused, future versions should explicitly discuss safeguards, dataset governance, and responsible use of synthetic identity-conditioned face generation.

\section*{Acknowledgment}

This study was financed in part by the \textit{Coordenação de Aperfeiçoamento de Pessoal de Nível Superior - Brasil~(CAPES)}, through the \textit{Programa de Excelência Acadêmica~(PROEX)} - Finance Code 001, and in part by the \textit{Conselho Nacional de Desenvolvimento Científico e Tecnológico~(CNPq)} and \textit{Fundação Araucária} under grant \#078/2026. The authors also thank the Federal Institute of Mato Grosso (IFMT), Pontes e Lacerda Campus, for supporting Bernardo Biesseck.

\bibliographystyle{IEEEtran}
\bibliography{references_arc2face_lcm}

@inproceedings{ho2020denoising,
  title={Denoising Diffusion Probabilistic Models},
  author={Ho, Jonathan and Jain, Ajay and Abbeel, Pieter},
  booktitle={Advances in Neural Information Processing Systems (NeurIPS)},
  year={2020}
}

@inproceedings{song2020denoising,
  title={Denoising Diffusion Implicit Models},
  author={Song, Jiaming and Meng, Chenlin and Ermon, Stefano},
  booktitle={Int. Conf. on Learning Representations (ICML)},
  year={2021}
}

@inproceedings{rombach2022high,
  title={High-Resolution Image Synthesis with Latent Diffusion Models},
  author={Rombach, Robin and Blattmann, Andreas and Lorenz, Dominik and Esser, Patrick and Ommer, Bj{\"o}rn},
  booktitle={CVPR},
  year={2022}
}

@inproceedings{deng2019arcface,
  title={ArcFace: Additive Angular Margin Loss for Deep Face Recognition},
  author={Deng, Jiankang and Guo, Jia and Xue, Niannan and Zafeiriou, Stefanos},
  booktitle={CVPR},
  year={2019}
}

@inproceedings{zhu2021webface260m,
  title={WebFace260M: A Benchmark Unveiling the Power of Million-Scale Deep Face Recognition},
  author={Zhu, Zheng and Huang, Guan and Deng, Jiankang and Ye, Yun and Huang, Junjie and Chen, Xinze and Zhu, Jiagang and Yang, Tian and Lu, Jiwen and Du, Dalong and Zhou, Jie},
  booktitle={CVPR},
  year={2021}
}

@inproceedings{papantoniou2024arc2face,
  title={Arc2Face: A Foundation Model for ID-Consistent Human Faces},
  author={Papantoniou, Foivos Paraperas and Lattas, Alexandros and Moschoglou, Stylianos and Deng, Jiankang and Kainz, Bernhard and Zafeiriou, Stefanos},
  booktitle={European Conf. on Computer Vision (ECCV)},
  year={2024}
}

@inproceedings{song2023consistency,
  title={Consistency Models},
  author={Song, Yang and Dhariwal, Prafulla and Chen, Mark and Sutskever, Ilya},
  booktitle={Int. Conf. on Machine Learning (ICML)},
  year={2023}
}

@article{luo2023latent,
  title={Latent Consistency Models: Synthesizing High-Resolution Images with Few-Step Inference},
  author={Luo, Simian and Tan, Yiqin and Huang, Longbo and Li, Jian and Zhao, Hang},
  journal={arXiv preprint arXiv:2310.04378},
  year={2023}
}

@article{luo2023lcmlora,
  title={LCM-LoRA: A Universal Stable-Diffusion Acceleration Module},
  author={Luo, Simian and Tan, Yiqin and Patil, Suraj and Gu, Daniel and von Platen, Patrick and Passos, Apolin{\'a}rio and Huang, Longbo and Li, Jian and Zhao, Hang},
  journal={arXiv preprint arXiv:2311.05556},
  year={2023}
}

@inproceedings{hu2022lora,
  title={LoRA: Low-Rank Adaptation of Large Language Models},
  author={Hu, Edward J. and Shen, Yelong and Wallis, Phillip and Allen-Zhu, Zeyuan and Li, Yuanzhi and Wang, Shean and Wang, Lu and Chen, Weizhu},
  booktitle={Int. Conf. on Learning Representations (ICLR)},
  year={2022}
}

@article{shi2025expertgen,
  title={ExpertGen: Training-Free Expert Guidance for Controllable Text-to-Face Generation},
  author={Shi, Liang and Fu, Yun},
  journal={arXiv preprint arXiv:2505.17256},
  year={2025}
}

@inproceedings{li2025interlcm,
  title={{InterLCM}: Low-Quality Images as Intermediate States of Latent Consistency Models for Effective Blind Face Restoration},
  author={Li, Senmao and Wang, Kai and van de Weijer, Joost and Khan, Fahad Shahbaz and Guo, Chun-Le and Yang, Shiqi and Wang, Yaxing and Yang, Jian and Cheng, Ming-Ming},
  booktitle={Int. Conf. on Learning Representations (ICLR)},
  year={2025}
}

@inproceedings{heusel2017gans,
  title={{GANs} Trained by a Two Time-Scale Update Rule Converge to a Local Nash Equilibrium},
  author={Heusel, Martin and Ramsauer, Hubert and Unterthiner, Thomas and Nessler, Bernhard and Hochreiter, Sepp},
  booktitle={Ad. in Neural Information Processing Systems},
  year={2017}
}

@inproceedings{liu2015deep,
  title={Deep Learning Face Attributes in the Wild},
  author={Liu, Ziwei and Luo, Ping and Wang, Xiaogang and Tang, Xiaoou},
  booktitle={IEEE/CVF Int. Conf. on Computer Vision (ICCV)},
  year={2015}
}

@inproceedings{moschoglou2017agedb,
  title={AgeDB: The First Manually Collected, In-the-Wild Age Database},
  author={Moschoglou, Stylianos and Papaioannou, Athanasios and Sagonas, Christos and Deng, Jiankang and Kotsia, Irene and Zafeiriou, Stefanos},
  booktitle={CVPRW},
  year={2017}
}

@inproceedings{kim2025vigface,
  title={VIGFace: Virtual Identity Generation for Privacy-Free Face Recognition},
  author={Kim, Minsoo and Sagong, Min-Cheol and Nam, Gi Pyo and Cho, Junghyun and Kim, Ig-Jae},
  booktitle={ICCV},  
  year={2025}
}

@inproceedings{kim2023dcface,
  title={DCFace: Synthetic Face Generation with Dual Condition Diffusion Model},
  author={Kim, Minchul and Liu, Feng and Jain, Anil and Liu, Xiaoming},
  booktitle={IEEE/CVF Conf. on Computer Vision and Pattern Recognition (CVPR)},
  year={2023}
}

@inproceedings{wu2025vec2face,
  title={Vec2Face: Scaling face dataset generation with loosely constrained vectors},
  author={Wu, Haiyu and Singh, Jaskirat and Tian, Sicong and Zheng, Liang and Bowyer, Kevin W},
  booktitle={Int. Conf. on Learning Representations (ICLR)},
  year={2025}
}

@inproceedings{Boutros2023IDiffFace,
    author    = {Fadi Boutros and Jonas Henry Grebe  and Arjan Kuijper and Naser Damer},
    title     = {IDiff-Face: Synthetic-based Face Recognition through Fizzy Identity-conditioned Diffusion Models},
booktitle = {IEEE/CVF Int. Conf. on Computer Vision (ICCV)},
    month     = {October},
    year      = {2023}
}

@inproceedings{He_2016_CVPR,
  author    = {He, Kaiming and Zhang, Xiangyu and Ren, Shaoqing and Sun, Jian},
  title     = {Deep Residual Learning for Image Recognition},
  booktitle = {IEEE/CVF Conf. on Computer Vision and Pattern Recognition (CVPR)},  
  year      = {2016}
}

@InProceedings{pmlr_v139_radford21a,
  title = 	 {Learning Transferable Visual Models From Natural Language Supervision},
  author =       {Radford, Alec and Kim, Jong Wook and Hallacy, Chris and Ramesh, Aditya and Goh, Gabriel and Agarwal, Sandhini and Sastry, Girish and Askell, Amanda and Mishkin, Pamela and Clark, Jack and Krueger, Gretchen and Sutskever, Ilya},
  booktitle = 	 {Int. Conf. on Machine Learning (ICML)},
  year = 	 {2021}
}

@inproceedings{ronneberger2015unet,
  title={U-net: Convolutional networks for biomedical image segmentation},
  author={Ronneberger, Olaf and Fischer, Philipp and Brox, Thomas},
  booktitle={MICCAI},
  year={2015},
}

@inproceedings{szegedy2016rethinking,
  title={Rethinking the Inception Architecture for Computer Vision},
  author={Szegedy, Christian and Vanhoucke, Vincent and Ioffe, Sergey and Shlens, Jonathon and Wojna, Zbigniew},
  booktitle = {CVPR},  
  year={2016}
}

\end{document}